\documentclass[times,twocolumn,final,authoryear]{elsarticle}

\usepackage{framed,multirow}

\usepackage{amssymb}
\usepackage{latexsym}
\usepackage{amsmath}
\usepackage{booktabs}
\usepackage{color}

\usepackage{url}
\usepackage{xcolor}
\definecolor{green1}{rgb}{0,.6,0}
\definecolor{purple1}{rgb}{.4,0,.8}
\definecolor{red1}{rgb}{.8,0,0}
\definecolor{blue1}{rgb}{0,0,1}

\usepackage{hyperref}
\hypersetup{
   colorlinks=true
   linkcolor=blue
   citecolor=green
}

\journal{Computer Vision and Image Understanding}

\begin{document}

\ifpreprint
  \setcounter{page}{1}
\else
  \setcounter{page}{1}
\fi

\begin{frontmatter}

\title{Self-Guiding Multimodal LSTM - when we do not have a perfect training dataset for image captioning}

\author[gc]{Yang Xian\fnref{fn1}}
\ead{yxian@gradcenter.cuny.edu}

\author[gc,ccny]{Yingli Tian}
\ead{ytian@ccny.cuny.edu}

\fntext[fn1]{The paper is under consideration at Computer Vision and Image Understanding.}

\address[gc]{The Department of Computer Science, The Graduate Center, The City University of New York, New York, NY, 10016}
\address[ccny]{The Department of Electrical Engineering, The City College, The City University of New York, New York, NY, 10031}

\begin{abstract}
In this paper, a self-guiding multimodal LSTM (sg-LSTM) image captioning model is proposed to handle uncontrolled imbalanced real-world image-sentence dataset. We collect FlickrNYC dataset from Flickr as our testbed with $306,165$ images and the original text descriptions uploaded by the users are utilized as the ground truth for training. Descriptions in FlickrNYC dataset vary dramatically ranging from short term-descriptions to long paragraph-descriptions and can describe any visual aspects, or even refer to objects that are not depicted. To deal with the imbalanced and noisy situation and to fully explore the dataset itself, we propose a novel guiding textual feature extracted utilizing a multimodal LSTM (m-LSTM) model. Training of m-LSTM is based on the portion of data in which the image content and the corresponding descriptions are strongly bonded. Afterwards, during the training of sg-LSTM on the rest training data, this guiding information serves as additional input to the network along with the image representations and the ground-truth descriptions. By integrating these input components into a multimodal block, we aim to form a training scheme with the textual information tightly coupled with the image content. The experimental results demonstrate that the proposed sg-LSTM model outperforms the traditional state-of-the-art multimodal RNN captioning framework in successfully describing the key components of the input images.
\end{abstract}

\end{frontmatter}


\section{Introduction}
\label{sec1}

In the recent popularized language-vision community, image captioning has been an important task. It involves generating a textual description that describes an image by analyzing its visual content. Automatic image captioning is able to assist solving computer vision challenges including image retrieval, image understanding, object recognition, navigation for the blind, and many others.

\begin{figure}
\begin{center}
\includegraphics[scale=0.5]{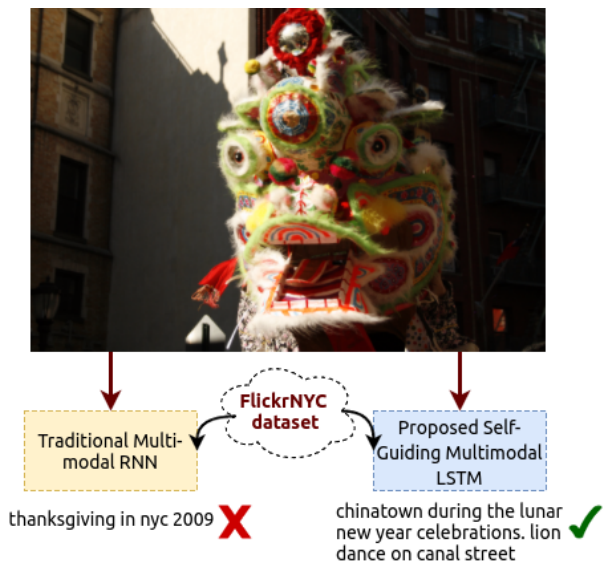}
\end{center}
   \caption{Example of the description generated by the proposed sg-LSTM image captioning framework compared with the result generated by the traditional multimodal RNN. Both frameworks are trained with FlickrNYC - a new dataset proposed in this paper. }
\label{fig:cover}
\end{figure}

Although image captioning is a natural task for human beings, it remains challenging from a computer vision point of view especially due to the fact that the task itself is ambiguous. There are countless ways to describe one input image, from high-level descriptions to explanations in details, while all are semantically correct. The fundamental cause is that in principle, descriptions of an image can talk about any visual aspects in it varying from object attributes to scene features, or even refer to objects that are not depicted and the hidden interaction or connection that requires common sense knowledge to analyze \citep{Bernardi2016}. 

In general, image captioning is a data-driven task. Descriptions for query images are normally defined by the training data. Therefore, it is not uncommon to see the birth of a new dataset for a new task. Recently, Amazon Mechanical Turk (AMT) is involved in more and more dataset description generation process. Different sets of descriptions may be generated depending on the instructions provided to fit a specific captioning task. Since it is an expensive process, majority of the image captioning frameworks focus on exploring existing datasets which tend to provide a sentence description embedded with the objects, attributes, and the reactions with the scene in the image. Some other frameworks tackle the problem from a different angle, such as unambiguous descriptions \citep{Mao2016}, image stream descriptions \citep{crcn}, etc. 

In this paper, we work with FlickrNYC - an image-sentence dataset collected directly from Flickr. The original descriptions provided by the users are utilized as the training data. Flickr data has been widely used in the dataset collection \citep{SBU1M,Flickr8K,Flickr30K} due to its availability of billions of images. However, the descriptions provided by the users are rarely used for captioning purpose directly due to several characteristics of the Flickr text data: $1$) Lengths of the descriptions vary dramatically for each image. While some users talk in paragraphs about the details including the possible background that is not directly related to the image, others may just describe in a few words indicating the location or the date information. $2$) Users may input descriptions for an album instead of a photo. Therefore, we may have multiple images that look visually different with the same description. $3$) Unlike the labeling process performed by AMT workers, the content of the descriptions is not strictly controlled semantically or syntactically. Foreign languages exist along with personal information including copyright statement, camera information, and links to personal social media accounts. Existing natural language processing (NLP) tools provide a limited solution in preparing the training data. In the end, it becomes tedious to set filtering criteria or use regular expressions to generate the `perfect' training dataset.  

\begin{figure*}
\begin{center}
\includegraphics[scale=0.53]{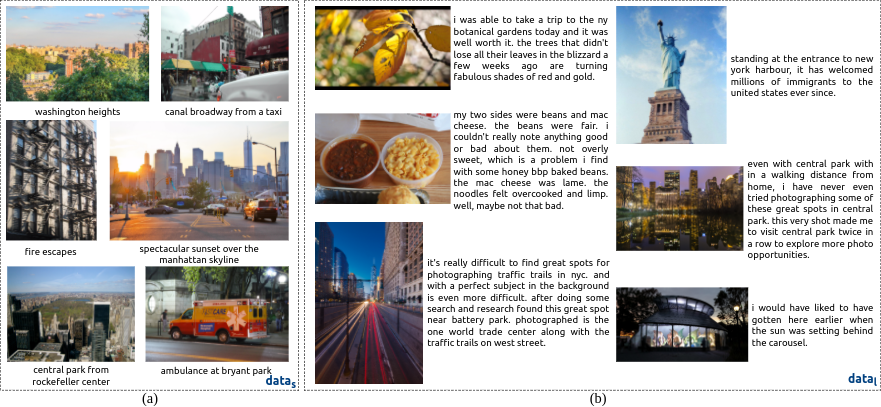}
\end{center}
   \caption{Sample images and the corresponding descriptions in FlickrNYC dataset. (a) Examples from $data_s$ in which images are with short descriptions. (b) Examples from $data_l$ in which images are with long descriptions. }
\label{fig:dataset_example}
\end{figure*}

However, despite all the characteristics listed, Flickr data meets the criterion for image captioning task - it comes from millions of users who can describe anything related to the images they upload. And more importantly, it is a real-world uncontrolled valuable resource. In this paper, we use `new york city' as our test case, i.e., `new york city' is employed as the keyword for the query process, to build FlickrNYC dataset\footnote{We will release the FlickrNYC dataset to public.}. As observed in FlickrNYC, descriptions in shorter lengths are more strongly correlated to the image content and are mainly related to the locations, events, or activities. They occur more repetitively compared with longer descriptions, e.g., a user may uploaded several images related to a walk in central park and they all have the description as `central park'. On the other hand, descriptions in longer sentences or paragraphs reveal more syntactical details, but may provide concepts that are more implicitly related to the images and have a weaker or no correlation to the image content. Examples in FlickrNYC can be found in Fig.~\ref{fig:dataset_example}.

In the proposed framework, a self-guiding multimodal long short-term memory (sg-LSTM) framework is presented to leverage between two portions of the data: $data_s$ (images with shorter length of descriptions) and $data_l$ (images with longer length of descriptions). We aim to make use of the part of the dataset with more reliable information to guide the training process of the caption generation. As demonstrated in Fig.~\ref{fig:cover}, a direct training utilizing the state-of-the-art multimodal RNN captioning method fails to capture the core event revealed in the image due to the fact that, FlickrNYC is a noisy real-world dataset in which we have multiple images labeled as `thanksgiving' but are visually different. Moreover, thanksgiving celebration is more frequently seen than Chinese New Year celebration within the training dataset. However, the proposed framework manages to generate accurate description that is both semantically and syntactically correct. 

Contributions of the proposed framework are threefold:
\begin{itemize}
\item A novel image captioning framework is proposed to deal with the uncontrolled image-sentence dataset where descriptions could be strongly or weakly correlated to the image content and in arbitrary lengths. The self-guiding process looks into the learning process in a global way to balance the syntactic correctness and the semantic details revealed in the images. This scheme can be extended to handle other tasks when we have imperfect training data.
\item A new FlickrNYC dataset is introduced with $306,165$ images related to `new york city'. Light pre-processing combining basic NLP tools and regular expression filtering are performed to remove the personal information including copyright, camera info., URLs, social network accounts, etc. Different from the majority captioning datasets, descriptions in FlickrNYC come from the original Flickr users.
\item Experimental results demonstrate that the proposed guiding textual feature manages to provide additional text information which strongly correlates to the image content. Compared with the existing traditional multimodal RNN captioning framework, the self-guiding scheme is able to recover more accurate descriptions given an input image.
\end{itemize}

The rest of the paper is organized as follows: Related work is discussed in Sec.~\ref{sec2}. Sec.~\ref{sec3} provides a detailed description of the proposed image captioning framework based on self-guiding multimodal LSTM. The collected dataset including the experimental results are presented in Sec.~\ref{sec4}. Conclusions are drawn in Sec.~\ref{sec5}.

\section{Related Work}
\label{sec2}

\subsection{Image Captioning}

Based on the underlying models utilized, recent image captioning frameworks can be classified into three categories. The first group of approaches casts the problem as a retrieval problem in which description of a test image is generated by searching for similar images in a database. This group of models employs the visual space to measure the similarity during image search. Descriptions of these similar images are transferred to obtain the target description.  \cite{Yagcioglu2015} utilized the activations of the last layer of the Visual Geometry Group convolutional neural network (VGG-CNN) \citep{vgg} trained on ImageNet \citep{imagenet} to represent the image features. The description of the query image is represented as a weighted average of the distributed representations of the retrieved descriptions. Different from \cite{Yagcioglu2015}, \cite{Devlin2015} employed the n-gram overlap F-score between the descriptions to measure the description similarity. Other than the deep learning based approaches, traditional machine learning techniques are also utilized in this task \citep{Verma2017}. 

The second group of methods adopts pre-defined sentence templates to generate image descriptions. The missing components in the sentence structures are filled based on image understanding of the objects, attributes and the correlations between objects and the scene. \cite{Elliott2013} proposed a sentence generation model which parses a query image into a visual dependency representation (VDR) which then traversed to fill the missing slots in the templates. More linguistically sophisticated approaches \citep{Mitchell2012, Kuznetsova2014, Ortiz2015} were proposed to tackle the sentence generation. 

The third group of approaches integrates image understanding and natural language generation into a unified pipeline. In general, image content in terms of objects, actions, and attributes is represented based on a set of visual features. Later, this content information is utilized to drive a language generation system, e.g., a recurrent neural network (RNN), to output the image descriptions\citep{You2016,SCN}. Some frameworks model image and text jointly into a multimodal space where later the joint representation space is used to perform cross-modal retrieval based on a query image. \cite{Karpathy2015} presented an alignment model which uses a structured object to align the two modalities (i.e., CNN over image regions and bidirectional RNN over sentences) through a multimodal embedding. An encoder-decoder framework is presented by \cite{Kiros2015} utilizing a joint multimodal space in which the LSTM is a big success. Another represented work in this category is the m-RNN model \citep{mrnn} in which a multimodal component is introduced to explicitly connect the language model and the vision model by a one-layer representation. 

With image captioning being a thriving topic, it is driven by the technical trials and improvements in both computer vision and NLP, and also importantly, the availability of relevant datasets. Other than the traditional image captioning task, efforts have been made to special captioning tasks. \cite{Mao2015} modified m-RNN to address the task of learning novel visual concepts. \cite{DCC} incorporated unpaired image data with labeling and unpaired text data to address the concept limitations in the image-sentence paired dataset. Similarly, \cite{Venugopalan2017} proposed the Novel Object Captioner (NOC) to describe object categories that are not present in the existing image-sentence datasets. `Referring expression' was explored \citep{referitgame, Mao2016} to generate unambiguous descriptions. \cite{crcn} presented a coherence recurrent convolutional network (CRCN) to describe an image stream in a storytelling manner utilizing blog data. Later, the authors brought up the personalized image captioning framework counting in users' vocabulary in previous documents \citep{Park2017}. A fill-in-the-blank image captioning task was introduced by \cite{Sun2017}.

The proposed sg-LSTM captioning framework falls into the third category. A multimodal component is utilized to connect the visual and the textual spaces. Different from the existing methods, a novel guiding textual feature is proposed to emphasize the correlation between the description and the image content. The guiding text is extracted through a separate m-LSTM model and serves as an additional input to the multimodal component.

\subsection{Datasets}

Due to the rising interest in image captioning task, a number of datasets have been brought up varying in sizes, formats of the descriptions, and the collection process. One of the earliest benchmark datasets - Pascal1K \citep{Pascal1K} was proposed which consists of $1,000$ images selected from Pascal $2008$ object recognition dataset \citep{Everingham2010}. Each image is associated with five sentence descriptions generated by AMT.

Later, based on Pascal $2010$ action recognition dataset, \cite{Elliott2013} introduced the Visual and Linguistic Treebank (VLT2K)  with $2,424$ images. AMT is again utilized with specific instructions to generate three, two-sentence descriptions for each image. Object annotation is available for a small subset of the images and VDRs are created manually for these images.

The Flickr8K \citep{Flickr8K} and Flickr30K \citep{Flickr30K} find their roots on images from Flickr. Although the images are collected based on user queries for specific objects or actions, the descriptions are generated in a manner similar to Pascal1K dataset where AMT workers provide five captions for each image. The original titles or descriptions from Flickr are not directly utilized to generate the captions in these two datasets. On the other hand, user-provided descriptions are employed in SBU1M \citep{SBU1M} which contains approximately one million captioned images from Flickr. Strict filtering is applied that the downloaded image should contain at least one noun and one verb on predefined control lists.

The MS COCO dataset \citep{mscoco} is widely used recently for image captioning evaluation with $123,287$ images accompanied by five descriptions per image. Extensions of MS COCO dataset are available to meet specific needs of various tasks, e.g., question answering \citep{Antol2015}, unambiguous descriptions \citep{Mao2016}, and text detection and recognition \citep{Veit2016}. The D\'{e}j\`{a} image captions dataset \citep{Deja} makes use of the existing web data without additional human efforts. It consists $4$ million images with $180$K unique captions where lemmatization and stop word removal are employed to normalize the captions and create a corpus of near-identical texts.

\begin{figure*}
\begin{center}
\includegraphics[scale=0.3]{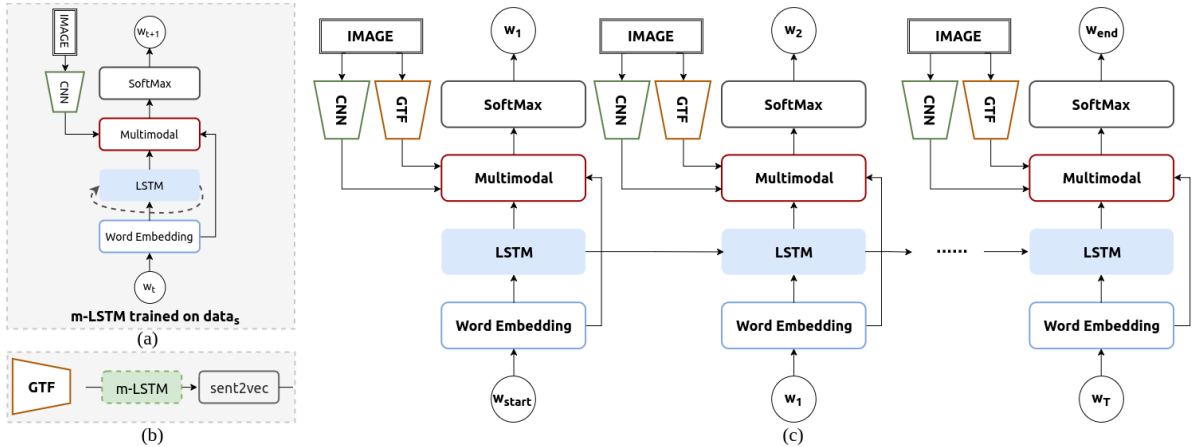}
\end{center}
   \caption{Systematic flowchart of the proposed self-guiding multimodal LSTM (sg-LSTM) captioning framework. (a) Basic multimodal LSTM (m-LSTM) captioning framework trained on a subset of the FlickrNYC dataset with short descriptions (i.e., $data_s$). $w_t$ denotes the $t$-th word in a sentence with words ranging from $w_1$ to $w_T$. A start sign $w_{start}$ and an end sign $w_{end}$ are added to all training instances in both $data_s$ and $data_l$. (b) Guiding text feature (GTF) extraction: to extract the text feature for self-guiding, we generate the descriptions utilizing m-LSTM followed by the sentence vectorizer. (c) Illustration of the sg-LSTM architecture. Compared with m-LSTM, an additional textual feature is fed to the multimodal block which encodes the language information connected to the image content. Figure is best viewed in color.}
\label{fig:flowchart}
\end{figure*}

Although various datasets have been collected recently, expensive human label under specific instructions or strict filtering is often required especially for large datasets. However, as mentioned, image descriptions should come in different degrees of abstraction, i.e., descriptions could be abstract as in several words or a short term, or as a detailed paragraph in a storytelling way. FlickrNYC dataset is collected through the real-world user data, which meets the requirements for image description naturally and the additional human efforts for the dataset generation is minimized without NLP-based normalization. 

\section{Deep Captioning via Self-Guiding Multimodal LSTM}
\label{sec3}

The successful combination of CNN and RNN, especially LSTM, has been widely experimented in image captioning and related tasks. However, as observed by \cite{gLSTM}, the generated sentence sometimes is weakly coupled to the provided image but is strongly correlated to the high frequency sentences in the training dataset. This is due to the fact that the generated sentence is ``drifted away'' during the sequence prediction process. This problem exists especially for long sentences where the generation is carried out ``almost blindly towards the end of the sentence''. To address this issue, alternative extensions have been proposed by adding attention mechanism \citep{Xu2015} and modifying the LSTM cell \citep{gLSTM}. However, it is still challenging with an uncontrolled dataset with descriptions in arbitrary lengths and abstraction levels.

In this section, we first introduce the basic multimodal LSTM (m-LSTM) image captioning framework which fuses the information of the input sentences and the corresponding image features in the multimodal component. It works effectively when the two input sources are strongly bonded. However, when this is not the case, it is difficult to maintain the correlation as the sentence generation goes on especially when the training dataset is not ideal for image captioning task. 

As observed in FlickrNYC dataset, descriptions in shorter sentences tend to have a stronger bond with the image content compared with longer descriptions. Although they may not be syntactically sound to form a sentence, these short descriptions tend to accurately describe the locations, activities, objects, or events, as the images were taken. Some examples can be found in Fig.~\ref{fig:dataset_example}(a) where core information in these images are conveyed in the corresponding descriptions. On the other hand, long descriptions are valuable as the users may state their feelings, reasoning, personal experiences, or objects that are not depicted in the image. As shown in Fig.~\ref{fig:dataset_example}(b), these sentences are difficult to reproduce by the AMT workers even with instructions. However, some descriptions may not be strongly bonded with the visual content.

In order to generate image captions with adequate details related to the image content, we separate the data based on the different characteristics revealed. FlickNYC is divided into two subsets, $data_s$ with descriptions in short sentences or terms, and $data_l$ with descriptions in long sentences or paragraphs (the length is measured in the number of words). We start by training a m-LSTM captioning model based on $data_s$. This captioning model aims to extract the key textual information provided an input image. This key information is later utilized to guide the training of sg-LSTM based on $data_l$ to better link the description to the image content. This guiding information is represented through a sentence vectorizer and fed as another input to the multimodal component in sg-LSTM.

\subsection{Captioning with m-LSTM}

To train a caption model with $data_s$, we employ a variation of m-RNN \citep{mrnn} due to its elegance and simplicity. The gated recurrent unit is replaced with LSTM in the proposed m-LSTM model. The LSTM network \citep{LSTM} has been widely used to model temporal dynamics in sequences. Compared with the traditional RNN, it better addresses the issue of exploding and vanishing gradients. The basic LSTM block consists of a memory cell which stores the state over time and the gates which control how to update the state of the cell. 

As illustrated in Fig.~\ref{fig:flowchart}(a), m-LSTM is composed of a word embedding layer, an LSTM layer, a multimodal layer, and a softmax layer. It takes the training images and the corresponding descriptions as inputs. Each word in the sentences is encoded with one-hot representation before being fed to m-LSTM training. The word embedding layer aims to map the one-hot vector to a more compact representation as shown in Eq.~\ref{eq:1}. Same as \cite{mrnn}, we randomly initialize the embedding layer and learn $W_e$ during training.

\begin{equation}
\label{eq:1}
e_t = W_e \cdot w_t ,
\end{equation}
where $w_t$ stands for the one-hot representation of word at step $t$. $W_e$ is the mapping weight between the one-hot representation and the word embedding representation $e_t$.

There are many LSTM variants. In the proposed m-LSTM model and later in sg-LSTM, we adopt LSTM with peepholes \citep{LSTMP} where the memory cell and gates within an LSTM block are defined as:

\begin{equation}
i_t = \sigma(W_{ic}C_{t-1}+W_{ih}h_{t-1}+W_{ie}e_t+b_i) ,
\end{equation}
\begin{equation}
f_t = \sigma(W_{fc}C_{t-1}+W_{fh}h_{t-1}+W_{fe}e_t+b_f) ,
\end{equation}
\begin{equation}
o_t = \sigma(W_{oc}C_t+W_{oh}h_{t-1}+W_{oe}e_t+b_o) ,
\end{equation}
\begin{equation}
C_t = f_t \odot C_{t-1}+i_t \odot g_1(W_{ch}h_{t-1}+W_{ce}e_t+b_c) ,
\end{equation}
\begin{equation}
h_t = o_t \odot C_t ,
\end{equation}
in which $\odot$ denotes the element-wise product. $\sigma(\cdot)$ is the sigmoid nonlinearity-introduce function. $g_1(\cdot)$ is the basic hyperbolic tangent function. $i_t$, $o_t$, $f_t$, $C_t$, and $h_t$ represent the state values of the input gate, output gate, forget gate, cell state, and hidden state, respectively. $W_{[\cdot][\cdot]}$ and $b_{[\cdot]}$ denote the weight matrices and bias vectors for corresponding gates and states.

The word sequence is fed to the LSTM network by iterating the recurrence connection as shown in Fig.~\ref{fig:flowchart}(a). Inception v$3$ \citep{inceptionv3} is used to extract the image features. They are connected with the language inputs though a multimodal component. The multimodal part fuses the language information represented as the dense word embedding and the LSTM activation with the image information represented using CNN as shown below: 

\begin{equation}
mm_t = g_2(W_i\cdot I+W_d\cdot e_t+W_l\cdot \text{LSTM}_t) ,
\end{equation}
where $g_2(\cdot)$ is the element-wise scaled hyperbolic tangent function \citep{LeCun2012} which leads to a faster training process than the basic hyperbolic tangent function. $W_i$, $W_d$ and $W_l$ indicate the mapping weights to learn during training.

The m-LSTM model is learnt utilizing a log-likelihood cost function based on perplexity introduced by \cite{mrnn}:
\begin{equation}
C_{\text{m-LSTM}} = \frac{1}{N_w}\displaystyle\sum_{i=1}^{N_s}L_i \cdot \log_2 \mathcal{P}(w_{1:L_i}^{(i)}|~I^{(i)}) + \lambda_{\theta}\cdot \|\theta\|_{2}^{2} ,
\end{equation}
where $\mathcal{P}(\cdot)$ stands for the perplexity of a sentence given the image. $N_w$ and $N_s$ represent the number of words and the number of sentences in the training set. $L_i$ is the length of the $i$-th sentence, and $\theta$ denotes the model parameters. 

\subsection{Captioning with sg-LSTM}

In this subsection, we describe in detail the training of sg-LSTM with $data_l$. As mentioned, for some training instances in $data_l$, there is not a strong connection between the textual description and the image content. In other words, additional textual features are needed during training. Therefore, as presented in Fig.~\ref{fig:flowchart}(b), a guiding textual feature (GTF) extractor is proposed which connects a m-LSTM captioning model trained on $data_s$ to a sentence vectorizer. This guidance feature aims to provide additional textual information for each training instance in $data_l$, which tends to emphasize the correlation between the textual and the visual domains. Compared with the basic m-LSTM architecture, sg-LSTM carries additional information in the multimodal component. Same as the image feature, the guiding textual features are fed into the multimodal component on each timestep as auxiliary information. This additional textual feature implicitly encodes the semantic information related to the image, such as location, activity, etc.

The sg-LSTM architecture is composed of four layers in each timestep similar to m-LSTM. The embedding layer encodes the one-hot word representation into a dense word representation. The weights in the embedding layer are learnt from the training data aiming at encoding the syntactic and semantic meaning of the words. The word representation after the embedding layer serves as the input to the LSTM layer. Same as m-LSTM, we adopt a basic LSTM block with peepholes. After this layer, a multimodal layer is set to connect the CNN-based image feature, the dense word representation, the recurrent layer output, and the proposed guiding texture feature. The activations of these four inputs are mapped to the same multimodal feature space as the activation of the multimodal layer:

\begin{equation}
mm^{'}_{t} = g_2(W_{i}^{'}\cdot I+W_{d}^{'}\cdot e_t+W_{l}^{'}\cdot  \text{LSTM}_t+W_{t}^{'}\cdot T) ,
\end{equation}
where $W_{i}^{'}$, $W_{d}^{'}$, $W_{l}^{'}$, and $W_{t}^{'}$ represent the corresponding weighting matrices.\\

\noindent\textbf{Extraction of Guiding Textual Features:}

\begin{figure*}[t]
\begin{center}
\includegraphics[scale=0.52]{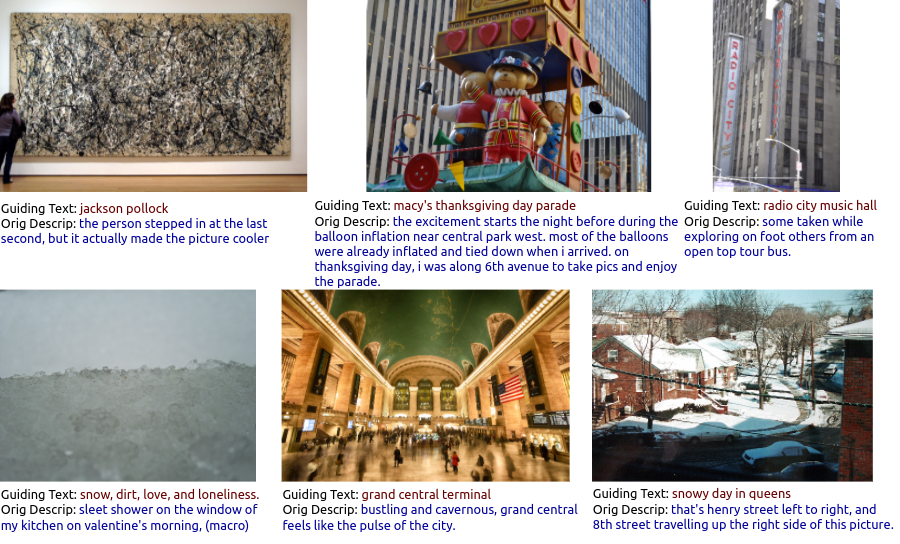}
\end{center}
   \caption{Examples of the guiding text (marked in red) generated by m-LSTM compared with the original descriptions (marked in blue) provided by the Flickr users. The guiding text provides supplementary information that is strongly related to the image content. Figure is best viewed in color. }
\label{fig:GTF}
\end{figure*}

To generate the guiding textual feature for a certain image, we first utilize m-LSTM trained on $data_s$ to output a short description (i.e., the raw sentence for the guiding textual feature). Beam search is adopted in the process to avoid the exhaustive search in the exponential search space. It is widely used in RNN-based captioning models \citep{mrnn,crcn,gLSTM} due to its efficiency and effectiveness. The top $1$ ranked sentence is selected for further vectorization. Fig.~\ref{fig:GTF} presents examples of the guiding texts generated by m-LSTM trained on $data_s$. The images are from $data_l$ and therefore, the original descriptions are relatively long. As demonstrated, the guiding texts either provide core information that is not conveyed in the original descriptions, e.g., authorship info (jackson pollock), landmark name (radio city music hall), and season info (snowy day), or emphasize the key image content buried in long sentences, e.g., event (macy's thanksgiving day parade), and location (grand central terminal). We also observe some interesting results that reveal some underneath feelings of the images themselves, e.g., `snow, dirt, love, and loneliness'.

A group of sentence vectorizers are investigated to vectorize the sentence or term generated by m-LSTM. In general, we adopt the word$2$vec with fusion scheme, i.e., each word in the sentence is vectorized and then these word vectors are combined to produce the final output. Three word2vec schemes are experimented:

\begin{itemize}
\item word$2$vec-GloVe: we adopt pre-trained GloVe \citep{glove}, i.e., Global Vectors for word representation, as the word vectorizer. The word vectors are trained through aggregated global word-word co-occurrence statistics from a corpus combining Wikipedia $2014$ and Gigaword $5$. We test two different feature dimensions, $50$ and $300$.
\item word$2$vec-NYC: compared with word$2$vec-GloVe, word$2$vec-NYC is a local word vectorizer trained on the textual data in FlickrNYC. This model is trained utilizing gensim \citep{gensim} and a $128$-dimensional vector is generated per word.
\item word$2$vec-short: a word embedding mapping is learnt when training m-LSTM on $data_s$. In this word vectorizer, the representation after the word embedding layer is employed directly as to map a word to a $1,024$-dimensional vector.
\end{itemize}

After representing each word in vector, two different fusing methods are investigated:

\begin{itemize}
\item Average: an average of all the word vectors in a sentence is calculated to obtain the final sentence vector.
\item TF-IDF: the word vectors are combined using term frequency-inverse document frequency (TF-IDF) weighting scheme to generate the final representation.
\end{itemize}

The various vectorization methods look into the mapping problem from different angles, utilizing a global corpus or a local dataset, and in different dimensionality. As later shown in Table~\ref{tab:results}, sg-LSTM based on word$2$vec-GloVe with TF-IDF weighting under feature dimension $50$ (denoted as sgLSTM-GloVe-tfidf-$50$) works the best among all the $8$ vectorization schemes (more details can be found in Sec.~\ref{sec4.3}).    \\

\noindent\textbf{Training sg-LSTM}

Same as m-LSTM, a log-likelihood cost function related to the perplexity is utilized for training sg-LSTM as shown in Eq. (8). Normalization regarding the number of words corrects the bias over shorter sentences during the caption generation process, and therefore, is suitable for FlickrNYC with images in various lengths.

\section{Experimental Results}
\label{sec4}

In this section, the effectiveness of the proposed self-guiding strategy is verified experimentally on FlickrNYC. We start by a deeper introduction of FlickrNYC dataset followed by the implementation details of the proposed system. Afterwards, experimental evaluation results are presented and analyzed. 

\subsection{FlickrNYC Dataset}
\label{sec4.1}

The FlickrNYC dataset is composed of $306,165$ images in total collected from Flickr with key word `new york city'. More specifically, Flickr search API is employed to crawl image-description data based on the key word, i.e., photos whose title, description, or tags contain `new york city' will be fetched. After capturing the images and their corresponding metadata, each image is accompanied with $1$ reference description provided by the original user. Images without valid descriptions are discarded. We perform a light pre-processing utilizing NLTK Toolbox \citep{NLTK}, textacy\footnote{\url{http://textacy.readthedocs.io/en/latest/index.html}}, and self-defined regular expressions, to remove unnecessary personal information (e.g., URLs, copyright declaration, camera information, personal social media accounts, advertisements, etc.). 

After the textual pre-processing, the dataset is divided based on the number of words in the descriptions. Images with descriptions shorter than $10$ form dataset $data_s$ with $165,374$ images for training and $1,000$ for testing. The rest $139,791$ images form $data_l$ in which $137,791$ is used for training, $1,000$ for validation and $1,000$ for
 testing. Table~\ref{tab:FlickrNYC} provides the statistics of distributions based on the description lengths in FlickrNYC. Sample images and the corresponding descriptions can be found in Fig.~\ref{fig:dataset_example}.

\begin{table}[h]
\centering
  \caption{Statistics of image distribution based on the description lengths in FlickrNYC dataset.}
  \label{tab:FlickrNYC}
  \resizebox{8.0cm}{!}{
  \begin{tabular}{l|ccccc}
    \toprule
    sentence length & $1-5$ & $6-10$ & $11-15$ & $16-25$ & $\geq{26}$\\
    num of instances & $94,180$ & $85,691$ & $45,291$ & $37,108$ & $43,895$ \\
  \bottomrule
\end{tabular}}
\end{table}

Different from the traditional way to create the vocabulary which removes all words that contain non-alphanumeric characters or even non-alphabetic characters, the vocabulary build-up process for FlickrNYC is tricky: 1) Since the dataset is based upon New York city in which multiple landmark names contain combinations of alphanumeric characters (e.g. `$5$th avenue' or informally `$5$ ave' in some descriptions), therefore, numeric and alphanumeric words should not be eliminated in the vocabulary. Moreover, words that contain or are connected by punctuations should also be considered, e.g., `Macy's', `it's', `let's', `sight-seeing', 'African-Americans', etc. 2) Although one image is accompanied by one description, the description is not restricted to one sentence. As observed, some descriptions can be long containing multiple sentences. To better model the continuity of a paragraph of sentences, punctuations such as `~,~', `~.~', `~!~', and `~?~' should be considered as part of the vocabulary list. 3) FlickrNYC utilizes uncontrolled real-world text data, which indicates that the usage of words can be informal. However, we find sometimes this informality is valuable since it reveals the emotions of the users, such as Emoticons (`~:-)~', `~:-P~', etc.) and exaggerated expressions (`soooo', `superrrr', etc.). Therefore, in order to keep all the information mentioned above, after tokenization and converted to lowercase, words that appear at least $3$ times in the training set are kept to create the vocabulary\footnote{If a word only contains alphabetic characters, we employ WordNet \citep{wordnet} to rule out typos and non-English words.}. The final vocabulary size is $22,230$.

\begin{table*}[t!]
\centering
  \caption{Numerical results of the proposed framework compared with other methods based on the testing images in $data_l$.}
  \label{tab:results}
  \resizebox{13cm}{!}{
  \begin{tabular}{l|ccccccc}
    \toprule
    & BLEU-1 & BLEU-2 & BLEU-3 & BLEU-4 & METEOR & ROUGE-L & CIDEr\\
    \midrule
    m-RNN \citep{mrnn} & $0.036$ & $0.019$ & $0.000$ & $0.000$ & $0.021$ & $0.084$ & $0.003$\\
    m-LSTM-long & $0.310$ & $0.257$ & $0.216$ & $0.169$ & $0.145$ & $0.244$ & $0.696$\\
    sgLSTM-NYC-ave & $0.237$ & $0.194$ & $0.160$ & $0.133$ & $0.122$ & $0.198$ & \textbf{1.270}\\
    sgLSTM-GloVe-tfidf-50 & \textbf{0.417} & \textbf{0.381} & \textbf{0.359} & \textbf{0.339} & \textbf{0.211} & \textbf{0.365} & $1.010$\\
    sgLSTM-GloVe-tfidf-300 & $0.281$ & $0.279$ & $0.278$ & $0.276$ & $0.154$ & $0.248$ & $0.177$\\
  \bottomrule
\end{tabular}}
\end{table*}

\subsection{Implementation Details} 
\label{sec4.2}

The proposed framework is built upon m-RNN\footnote{\url{https://github.com/mjhucla/TF-mRNN}} with TensorFlow \citep{tensorflow}. The inception v3 \citep{inceptionv3} pretrained on ImageNet \citep{imagenet} is used to compute CNN features as image representation. Feature dimension for this image representation is $2,048$. In both m-LSTM and sg-LSTM, the word embedding layer is with $1,024$ dimension. The LSTM layer and the multimodal layer are with $2,048$ dimensions. We assign $0.5$ dropout rate to all three layers.

Both m-LSTM and sg-LSTM models are trained with RMSProp optimizer \citep{RMSProp}. We apply the stochastic gradient descent (SGD) with mini-batches of $64$. The beam search size is set to be $3$. The top ranked sentence generated by m-LSTM based on training data in $data_s$ is utilized for guiding textual feature extraction. As mentioned, three word$2$vec schemes are tested, i.e., word$2$vec-GloVe, word$2$vec-NYC, and word$2$vec-short. Two different sets of pre-trained word vectors are tested for word$2$vec-GloVe with dimensions $50$ and $300$. Dimensions for word$2$vec-NYC and word$2$vec-short based representations are $128$ and $1,024$, respectively. 

\subsection{Experimental Evaluations}
\label{sec4.3}

In order to select the best vectorization scheme for the guiding textual feature, certain objective criterion is needed to evaluate each scheme. Popular evaluation metrics for image captioning tasks include BLEU \citep{BLEU} (BLEU@$1,2,3,4$), METEOR \citep{METEOR}, ROUGE-L \citep{ROUGE}, and CIDEr \citep{CIDEr}. However, none of the criteria listed is a perfect metric for the evaluation task in our case since the ground-truth descriptions in FlickrNYC dataset are noisy. An example can be found in the bottom left image in Fig.~\ref{fig:exp2} in which the original description is `december 6th'. On the other hand, the proposed sg-LSTM framework outputs description `boaters on the lake in central park near the bow bridge' which is a much better description compared with the original one given the image content. However, this superiority will not be reflected in the numerical metrics listed above.

\begin{figure*}
\begin{center}
\includegraphics[scale=0.42]{exp2.png}
\end{center}
   \caption{Descriptions generated by the proposed framework (marked in red) compared with m-RNN \citep{mrnn}, m-LSTM-long (m-LSTM trained on $data_l$), m-LSTM-full (m-LSTM trained on all training data) and the original descriptions (marked in blue) provided by the Flickr users. The guiding texts are also provided. To help with the evaluation, the ground-truth locations are marked in each image (usage of different colors is for the best contrast). Figure is best viewed in color.}
\label{fig:exp2}
\end{figure*}

\begin{figure*}
\begin{center}
\includegraphics[scale=0.48]{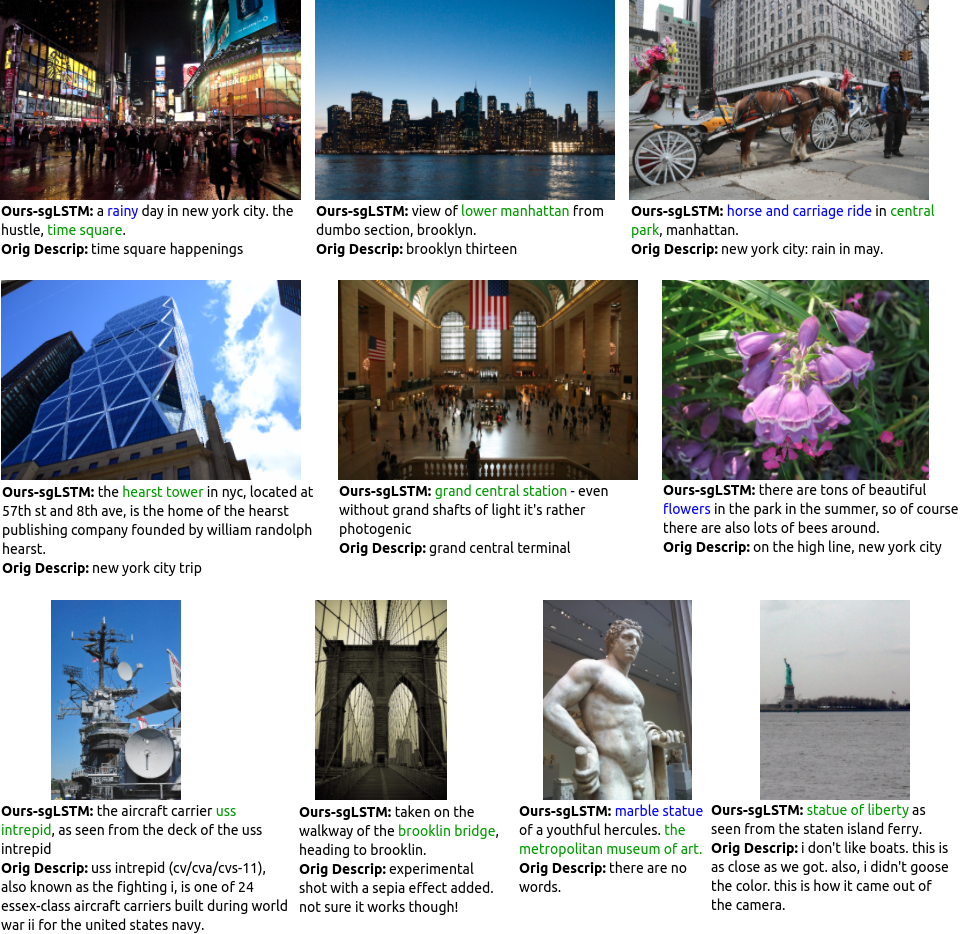}
\end{center}
   \caption{More results generated by the proposed framework compared with the original descriptions provided by the Flickr users. \textcolor{blue1}{Blue} indicates a precise description of the image content that does not even appear in the original descriptions. \textcolor{green1}{Green} shows a successful recovery of the landmarks. Figure is best viewed in color.}
\label{fig:exp3}
\end{figure*}

\begin{figure*}
\begin{center}
\includegraphics[scale=0.5]{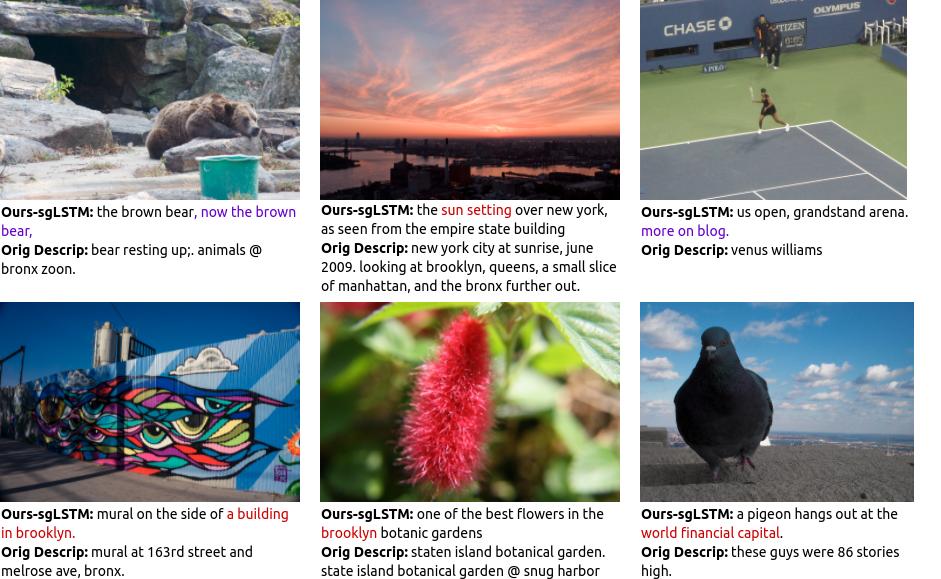}
\end{center}
   \caption{Results generated by the proposed framework that could be further improved. \textcolor{purple1}{Purple} indicates wrong or unrelated phrases. \textcolor{red1}{Red} shows a wrong location or activity based on the original descriptions provided by the users. However, for some cases, these locations or activities cannot be recovered solely based on the image content. Figure is best viewed in color.}
\label{fig:exp4}
\end{figure*}

Despite the challenges in evaluating the vectorization schemes in the proposed framework, there still exists a large portion of data in FlickrNYC which suits `perfectly' for captioning task. Therefore, a small validation dataset is separated from $data_l$ and utilized to evaluate the $8$ different vectorization schemes. Based on the experimental results, sgLSTM-GloVe-tfidf-50 achieves the best performance quantitatively. Thus we adopt sg-LSTM based on word$2$GloVe in TF-IDF weighting with dimension $50$ as the final setting and all the results reported in this paper are based on this setting unless stated otherwise.

Table~\ref{tab:results} presents the numerical results based on $1,000$ testing images in $data_l$. The proposed sg-LSTM framework is compared with m-RNN \citep{mrnn}, m-LSTM, and among different vectorization settings. The results of the top $3$ performers in the previous verification step are included in this table. m-LSTM-long represents the m-LSTM captioning model trained on $data_l$. As shown in Table~\ref{tab:results}, sgLSTM-GloVe-tfidf-50 gives the best performance numerically almost among all the evaluated methods, which is consistent with our observation in the verification step. 

The zero numbers shown in Table~\ref{tab:results} for m-RNN might be better explained by looking into the results in Fig.~\ref{fig:exp2}. A direct training over the whole dataset tends to put a preference into high frequency sentences in the training dataset, which may be unrelated to the test image itself. Therefore, when it comes to numerical evaluations, a total miss of the core concept in the image content leads to a low score. On the other hand, by integrating the guiding textual features into the training process, the proposed sg-LSTM model manages to generate accurate descriptions related to the image content, and sometimes, the generated descriptions are more meaningful than the original ones provided by the Flickr users as demonstrated in Fig.~\ref{fig:exp2}. 

Fig.~\ref{fig:exp3} provides more examples by comparing the results generated by the proposed framework with the original descriptions provided by the Flickr users. Our sg-LSTM model accurately generates descriptions that are closely related to the image content and successfully recovers key image features (e.g., weather, objects, activities) and the landmarks. In Fig.~\ref{fig:exp4}, several results are presented that could be further improved. Additional preprocessing steps could be performed before the training to remove terms such as `more on blog' as shown in the figure. NLP techniques can be applied in avoiding a repetitive pattern shown in the `bear' example. For certain cases, solely based on image content, it is difficult to generate accurate descriptions even for New Yorkers - as to differentiate between `sunset' and `sunrise', `brooklyn' and `bronx', or to decide the name of a certain building with little information - to that extent, it would be better to remove the ambiguous information that cannot be predicted based on the image content in the final description. 

\section{Conclusions}
\label{sec5}

In this paper, we have proposed a novel self-guiding multimodal LSTM captioning framework which targets at a more effective training over uncontrolled real-world dataset. A new FlickrNYC dataset is introduced as the testbed to verify the proposed self-guiding scheme. The portion of data, in which the textual description strongly correlates with the image content, is utilized to train a m-LSTM model to extract the textual features. Afterwards, the additional features are utilized to guide the training process of the caption generation based on the rest of the data. Experimental results demonstrate the effectiveness of the proposed framework in generating descriptions that are syntactically correct and semantically sound.

\bibliographystyle{model2-names}
\bibliography{acmmm_yang}

\end{document}